\documentclass[journal]{IEEEtran}
\usepackage{amsmath}
\usepackage{amssymb}
\usepackage{amsfonts}
\usepackage{xcolor}
\usepackage{booktabs}
\usepackage{cite}
\usepackage{footnote}
\usepackage{threeparttable}
\makesavenoteenv{tabular}
\usepackage{setspace}
\usepackage{graphicx}
\usepackage{multirow}
\usepackage{algorithm}
\usepackage{algorithmicx}
\usepackage{mathptmx}
\usepackage{algpseudocode}
\usepackage{diagbox}
\usepackage[hidelinks]{hyperref}
\usepackage{url}
\usepackage{float}
\usepackage{acronym}
\usepackage{amsmath,bm}

\begin{document}
\title{Weakly Supervised-Based Oversampling for High Imbalance and High Dimensionality Data Classification}
\author{Min Qian, Yan-Fu Li
	\thanks{The~authors~are~with~Department~of~Industrial~Engineering,~Tsinghua University,~Beijing,~P.R.~China.~(e-mail:~ liyanfu@tsinghua.edu.cn)}
}

\maketitle

\begin{abstract}
With the abundance of industrial datasets, imbalanced classification has become a common problem in several application domains. Oversampling is an effective method to solve imbalanced classification. One of the main challenges of the existing oversampling methods is to accurately label the new synthetic samples. Inaccurate labels of the synthetic samples would distort the distribution of the dataset and possibly worsen the classification performance. This paper introduces the idea of weakly supervised learning to handle the inaccurate labeling of synthetic samples caused by traditional oversampling methods. Graph semi-supervised SMOTE is developed to improve the credibility of the synthetic samples’ labels. In addition, we propose cost-sensitive neighborhood components analysis for high dimensionality datasets and bootstrapping based ensemble framework for high imbalance datasets. The proposed method has achieved good classification performance on 8 synthetic datasets and 3 real-world datasets, especially for high imbalance and high dimensionality problems. The average performances and robustness of the proposed method are better than the benchmark methods. 
\end{abstract}


\section{Introduction}
Classification is an important and general task of machine learning. One of the current challenges in this process is learning with a class imbalance dataset. Class imbalance means the number of instances from the positive class (minority) is much smaller than the number of instances of the negative class (majority). Since positive class occurs infrequently, they are likely predicted as rare occurrences, undiscovered or even ignored. Therefore, the misclassification rate of the class imbalance dataset is often much higher than that of the normal datasets \cite{RN1}. Most classification algorithms assuming a relatively balanced distribution have encountered serious difficulties to deal with imbalanced dataset.

Class imbalance often arises in many application domains such as software defect prediction \cite{RN2,song2018comprehensive}, medical diagnosis of a rare disease \cite{RN3,gan2020integrating} and fault diagnosis of industrial systems \cite{RN4,hu2018imbalance}, et al. The imbalanced classification problem has drawn a significant amount of research efforts from academia and industry in recent years.

Several specialized methods have been developed to address this problem. They can be divided into the following three main categories. (1) Resampling. Resampling methods are designed to reduce the imbalanced ratio. Two typical implementations include: oversampling which duplicates or generates new minority samples and undersampling which removes certain majority samples \cite{RN5,RN6}. (2) Cost-sensitive learning. For cost-sensitive learning, different types of misclassification have different penalty costs \cite{RN7}. The classifier assigns a higher cost to minority misclassification compared to majority misclassification, emphasizing any correct classification or misclassification regarding the majority class. (3) Ensemble learning. The ensemble methods will train several classifiers on balanced sub-datasets generated by resampling. Their evaluations are aggregated to produce the final classification decision \cite{RN9}. There are mainly two types of ensemble methods, i.e. Bagging and Boosting.

Generally, the datasets with imbalance ratio (IR) higher than 10:1 are regarded as the highly imbalanced datasets \cite{RN10}. In these datasets, over-fitting becomes difficult to avoid and the classification boundary of positive class is often unreliable. In this situation, oversampling methods have shown competitiveness to avoid overfitting and to improve the robustness of the classifier \cite{RN11}. The widely-used oversampling methods including randomly duplicating minority samples, the synthetic minority oversampling technique (SMOTE) \cite{RN5} and ADASYN \cite{he2008adasyn}. In recent years, some generative adversarial networks based oversamplings \cite{ali2019mfc} have been proposed. However, these oversampling methods, whether simple or complex, mainly focus on how to generate minority samples and ignore whether it is correct to label the generated samples directly as minority samples.

\subsection{Motivation}
One of the overlooked problems of the existing oversampling methods is how to properly label the synthetic data. Mainstream oversampling methods such as SMOTE will label all synthetic samples as positive class. This way is partially unreasonable, especially for those synthetic samples close to the classification boundary. This simple labeling method would distort the distribution of the original dataset, shift the classification boundary and result in a large number of misclassifications. Another issue is that the oversampling methods such as SMOTE usually generate samples using neighborhood information. In high-dimensional space, since the sample distribution becomes sparser, the neighborhood information becomes less reliable, resulting in poor quality of the generated samples. Hence, the performance of SMOTE on high-dimensional datasets is often poor. Finally, if the imbalanced ratio is high, oversampling method uses only a very limited number of positive samples to synthesize a large number of new samples. This would lead to a limited diversity of the synthetic samples, which is likely to cause over-fitting problems.

\subsection{Contribution}
The contributions of this study can be summarized as the following
three points:

\begin{itemize}
	\item We attempt to address the inaccurate labels of synthetic samples from the perspective of weakly supervised learning (WSL) \cite{RN12}. As a classic method of WSL, graph semi-supervised learning (GSSL) can effectively solve the problem with label noise in training set. In this article, we treat the synthetic samples as unlabeled data added to the training set, and then GSSL for labelling synthetic samples is developed to improve the credibility of synthetic samples’ labels.
	
	\item To better solve high-dimensional problems, we propose cost-sensitive neighborhood components analysis (CS-NCA) for dimensionality reduction. The imbalanced dataset is more separable after dimensionality reduction by CS-NCA.
	
	\item For highly imbalanced problems, bootstrapping based ensemble framework (BEF) is proposed for reducing the imbalanced ratio while avoiding the loss of negative class information. It effectively alleviates the overfitting problem in the training process.
\end{itemize}

We perform a set of experiments over 8 synthetic and 3 real-world datasets. The results show that our methods can achieve better performance than other benchmark methods on datasets of high dimensionality and high imbalance. We also observe that our approach is more robust to different datasets.

\section{Related Work}

\subsection{Defects of the exsiting oversampling method}
Assuming a binary classification problem, we have a imbalanced dataset $\mathbf{S}=\{\mathbf{X},\mathbf{Y}\}=\{(\mathbf{x}_1,y_1),(\mathbf{x}_2,y_2),...,(\mathbf{x}_n,y_n)\}$, labelling $\mathbf{Y}=(y_1,y_2,\ldots,y_n)\in\{{0,1}\}^n$, such that the entire dataset can be divided into negative (majority) class $\mathbf{S}_{0}$ and positive (minority) class $\mathbf{S}_{1}$, $\left|\mathbf{S}_{0}\right|\gg\left|\mathbf{S}_{1}\right|,\ \mathbf{S}=\mathbf{S}_{0}\cup\mathbf{S}_{1}$. When dealing with small and imbalanced datasets, appropriately capturing the joint probability function $P\left(\mathbf{X},\mathbf{Y}\right)$ might be unrealistic. Hence, most oversampling methods make use of the neighborhood information of the dataset to synthesize new samples.

In SMOTE, synthetic samples are generated by convex combination of some seed samples belonging to the positive class and labelled directly using the positive class label. The first seed sample $\mathbf{x}_{i}$ is chosen randomly from $\mathbf{S}_{1}$, and the other seeds are chosen as one of its $k$-nearest neighbors. $k$ is responsible for avoiding label inconsistencies and exploiting the local information of the dataset, but it also significantly limits the diversity of synthetic samples \cite{RN11}. Mathematically,

\begin{equation}
\mathbf{x}_{new}=\mathbf{x}_\mathbf{i}+\left({\hat{\mathbf{x}}}_\mathbf{i}-\mathbf{x}_\mathbf{i}\right)\times\delta,\ \ \delta\in U\left[0,1\right]
\end{equation}

where $\mathbf{x}_{new}$ is the synthetic sample, $\mathbf{x}_\mathbf{i}$ is a randomly selected positive sample, $k-NN(\mathbf{x}_\mathbf{i})$ is the set of the k-nearest neighbors of $\mathbf{x}_\mathbf{i}$ in $\mathbf{S}_\mathbf{1}$, ${\hat{\mathbf{x}}}_\mathbf{i}\in k-NN(\mathbf{x}_\mathbf{i})$, $\delta$ is a random number between 0 and 1. Note that SMOTE will label all synthetic samples as positive class, so $y_{new}\equiv1$.

\begin{figure}[t]
	\centering
	\includegraphics[width=1.0\columnwidth]{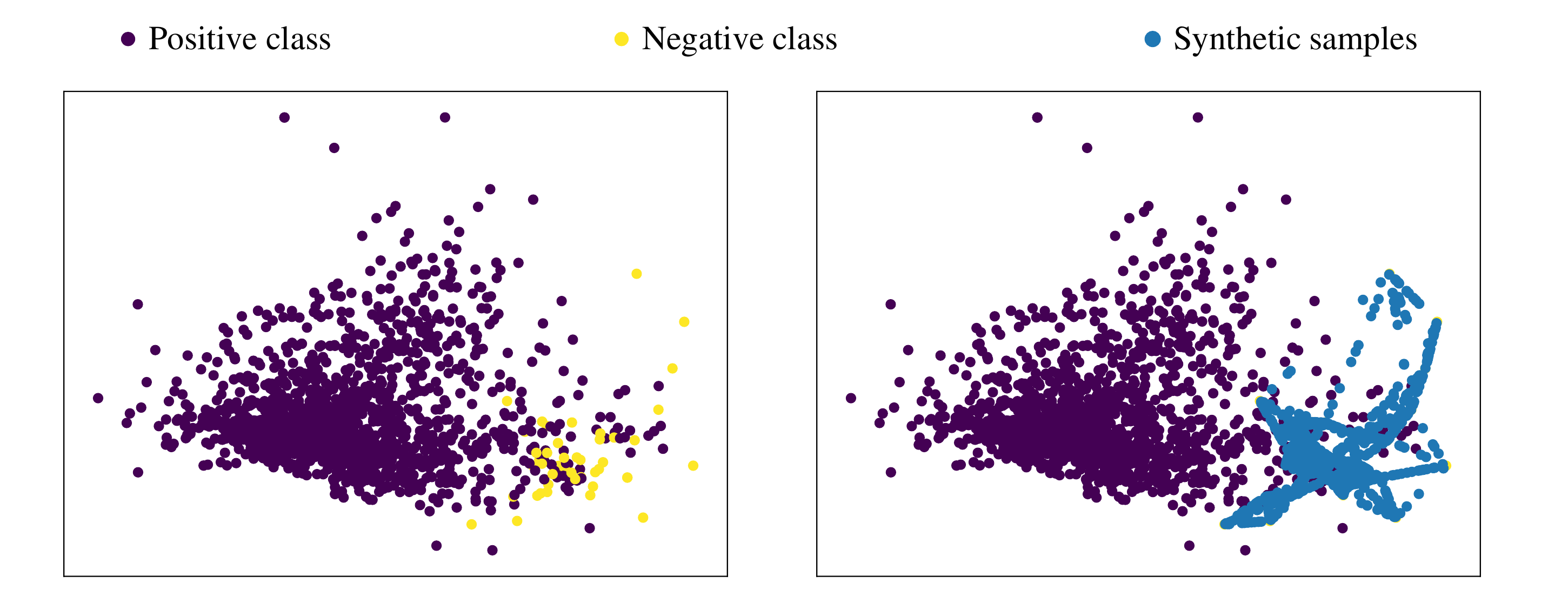}
	\caption{Example of SMOTE for yeast5 dataset, in which naively labelling synthetic data as minority class patterns might not be suitable.}
	\label{fig1}
\end{figure}

But there are still some limitations of the SMOTE. Fig.\ref{fig1} shows 2-dimension (after PCA) plot of yeast5 dataset from KEEL where the imbalanced ratio is 32.7:1 and the number of positive samples is 44 (left), some synthetic samples are created by SMOTE in order to balance the class distributions (right). This shows a representation of the main problem encountered using oversampling approach. Many synthetic samples are created in the region of the negative class or on the boundary between negative and positive class, especially when class overlapping occurs. If we simply label all these synthetic samples as positive class, it will obversely have a large impact on the classification accuracy. The borders of the negative class are severely damaged and the distribution of the dataset changed greatly. To solve this problem, we design a novel approach based on weakly supervised learning.

\subsection{Weakly supervised learning}
Due to the drawbacks of existing oversampling methods, we introduce the idea of weakly supervised learning in machine learning. The weakly supervised learning means that the supervision information of the provided data is incomplete, and the label of samples may be missing or inaccurate in classification problem. Typically, there are three types of weakly supervision \cite{RN12}, namely: incomplete supervision, inexact supervision and inaccurate supervision.

The problem mentioned above is one of the inaccurate supervisions. The true label of synthetic samples by oversampling may not always be positive. Inaccurate supervision concerns the situation in which the supervision information is not always ground-truth; in other words, some label information may suffer from errors. It is similar to the problem we have. An interesting recent scenario of inaccurate supervision occurs during crowdsourcing. For machine learning, crowdsourcing is commonly used as a cost-saving way to determine the labels for training data. In \cite{RN13}, an 'unsure' option is provided for crowdsourcing, such that labeling workers are not forced to give a label when they have low confidence. This option with theoretical support, helps improving the labeling reliability \cite{RN14}. The above research shows that the use of weakly supervised learning to solve imbalanced learning problems could be a feasible solution.

Perez-Ortiz et al. (2019) have used semi-supervised learning methods studied the effect of generating synthetic data by convex combination of patterns and the use of these as unsupervised information in a incomplete learning framework with S$^{3}$VM, thus avoiding the need to label synthetic examples \cite{RN11} . However, this method requires the dataset satisfying the clustering assumption \cite{RN12}, which is often not met in high imbalanced datasets.  

\section{Methods}

\subsection{Cost-sensitive neighborhood components analysis}
It is worth mentioning that oversampling methods generally rely on neighborhood information, and thus, they are less reliable in high-dimensional feature space because the identification of neighborhoods is usually less reliable when data points are relatively sparse. Thus dimension reduction is required at first \cite{RN16}. In this study, we propose a cost-sensitive neighbor components analysis (CS-NCA). It is a supervised dimensionality reduction algorithm based on metric learning \cite{RN17}. CS-NCA fully considers the characteristics of imbalanced datasets and introduces cost-sensitive ideas to make the classification boundary of the dataset more clearer. The purpose is to find a distance metric that maximizes the performance of the cost sensitive nearest neighbor classifier \cite{RN18}. To keep the distance non-negative and symmetric, the “metric matrix” $\mathbf{M}$ must be a positive semi-definite symmetric matrix. The distance metric finally learned can be expressed as

\begin{equation}
\mathbf{M}=\mathbf{P}\mathbf{P}^\mathbf{T}
\end{equation}

\begin{equation}
\begin{split}
d_M\left(\mathbf{x}_\mathbf{i},\mathbf{x}_\mathbf{j}\right) &=\left(\mathbf{x}_\mathbf{i}-\mathbf{x}_\mathbf{j}\right)^\mathbf{T}\mathbf{M}\left(\mathbf{x}_\mathbf{i}-\mathbf{x}_\mathbf{j}\right)\\
&=\left(\mathbf{P}^\mathbf{T}\mathbf{x}_\mathbf{i}-\mathbf{P}^\mathbf{T}\mathbf{x}_\mathbf{j}\right)^\mathbf{T}\left(\mathbf{P}^\mathbf{T}\mathbf{x}_\mathbf{i}-\mathbf{P}^\mathbf{T}\mathbf{x}_\mathbf{j}\right)
\end{split}
\end{equation}

By restricting $\mathbf{P}$ to be a nonsquare matrix of size $d\times D,d<D,d>rank(\mathbf{M})$. It can perform dimensionality reduction, where $D$ is the original data dimension and $d$ is the data dimension after dimensionality reduction.

The nearest neighbor classifier usually adopts the majority voting method. Each sample in the neighborhood casts one vote, and the samples outside the neighborhood cast zero vote. Here we replace it with the neighborhood probability voting method. For any sample $\mathbf{x}_\mathbf{j}$, the probability that it affects the classification result of $\mathbf{x}_\mathbf{i}$ is $p_{ij}$. Obviously, the influence of $\mathbf{x}_\mathbf{j}$ on $\mathbf{x}_\mathbf{i}$ decreases as their distance increases, where $\delta$ is the neighborhood distance threshold, it is set to eliminate the impact of the samples outside the neighborhood.

\begin{equation}
\label{equation4}
{r_{ij}} = \left\{ \begin{array}{l}
\exp ( - {d_M}({\mathbf{x}_i},{\mathbf{x}_j})),{d_M}({\mathbf{x}_i},{\mathbf{x}_j}) \le {\delta ^2}{\rm{\ and \ }}i \ne j\\
0,otherwise
\end{array} \right.
\end{equation}

\begin{equation}
p_{ij}=\frac{r_{ij}}{\sum_{l=1}^{n}r_{il}}
\end{equation}

Since the true data distribution cannot be obtained, the objective function can only attempt to maximize the accuracy of the leave-one-out (LOO) prediction on the training dataset. The LOO accuracy rate of $\mathbf{x}_\mathbf{i}$ is defined as the probability that it is correctly classified by all samples except itself:
\begin{equation}
P_i=\sum_{j\in\mathrm{\Omega}_i} p_{ij}
\end{equation}

Where $\mathrm{\Omega}_i$ represents the subscript set of samples belonging to the same class as $\mathbf{x}_\mathbf{i}$. For imbalanced dataset, we introduced the idea of cost-sensitive learning, making the feature space after dimensionality reduction more effective in distinguishing between positive and negative classes. Assign higher weight to the classification results of positive samples. The final objective function is defined as $Q$. $\mathbf{S}_\mathbf{0}$ is the negative class set, $\mathbf{S}_\mathbf{1}$ is the positive class set, $c>1$ is the classification weight of the positive sample, and here it is set as the imbalanced ratio of the training dataset.

\begin{equation}
\max_\mathbf{P}{\ Q=\sum_{i\in\mathbf{S}_\mathbf{0}}\sum_{j\in\mathrm{\Omega}_i} p_{ij}+c\sum_{i\in\mathbf{S}_\mathbf{1}}\sum_{j\in\mathrm{\Omega}_i} p_{ij}}
\end{equation}

Stochastic gradient descent is adopted to solve the above optimization problem. The derivation process of the target gradient is as follows:

\begin{equation}
\begin{split}
&\frac{\partial p_{ij}}{\partial\mathbf{P}}=\frac{1}{\left(\sum_{l} r_{il}\right)^2}\left(\frac{\partial r_{ij}}{\partial\mathbf{P}}\sum_{l} r_{il}-r_{ij}\sum_{l}\frac{\partial r_{il}}{\partial\mathbf{P}}\right)\\
&=-2p_{ij}\mathbf{P}\left[\left(\mathbf{x}_\mathbf{i}-\mathbf{x}_\mathbf{j}\right)\left(\mathbf{x}_\mathbf{i}-\mathbf{x}_\mathbf{j}\right)^\mathbf{T}-\sum_{l}{p_{il}\left(\mathbf{x}_\mathbf{i}-\mathbf{x}_\mathbf{l}\right)\left(\mathbf{x}_\mathbf{i}-\mathbf{x}_\mathbf{l}\right)^\mathbf{T}}\right]
\end{split}
\end{equation}

\begin{equation}
\begin{split}
\frac{\partial Q}{\partial\mathbf{P}}&=\sum_{i\in \mathbf{S}_0}\sum_{j\in\mathrm{\Omega}_i}\frac{\partial p_{ij}}{\partial\mathbf{P}}+c\sum_{i\in \mathbf{S}_1}\sum_{j\in\mathrm{\Omega}_i}\frac{\partial p_{ij}}{\partial\mathbf{P}}\\
&=2\sum_{i}\sum_{j}{(p_ip_{ij}-q_{ij})\left(\mathbf{P}\mathbf{x}_\mathbf{i}-{\mathbf{Px}}_\mathbf{j}\right)\left(\mathbf{x}_\mathbf{i}-\mathbf{x}_\mathbf{j}\right)^\mathbf{T}}
\end{split}
\end{equation}
where
\begin{equation}
{p_{i}} = \left\{ \begin{array}{l}
\sum\limits_{j \in {\Omega _i}} {{p_{ij}}} ,i \in {\mathbf{S}_0}\\
c\sum\limits_{j \in {\Omega _i}} {{p_{ij}}} ,i \in {\mathbf{S}_1}
\end{array} \right.
\end{equation}

\begin{equation}
{q_{ij}} = \left\{ \begin{array}{l}
{p_{ij}},i \in {\mathbf{S}_0}{\rm{\ and \ }}j \in {\Omega _i}\\
c \times {p_{ij}},i \in {\mathbf{S}_1}{\rm{\ and \ }}j \in {\Omega _i}\\
0,otherwise
\end{array} \right.
\end{equation}

Define the elements of matrix $\mathbf{H}$ as $h_{ij}=p_ip_{ij}-q_{ij}$, $g(\mathbf{H})$ is a diagonal matrix, the diagonal element is the column sum of the matrix $\mathbf{H}$, $g_i=\sum_{i=1}^{n}h_{ij}$. By the definition, we can easily get $g(\mathbf{H}^\mathbf{T})=\mathbf{0}$. The matrix form of the gradient formula is

\begin{equation}
\begin{split}
\frac{\partial Q}{\partial\mathbf{P}}&=2\sum_{i}\sum_{j}{h_{ij}\left(\mathbf{P}\mathbf{x}_\mathbf{i}-{\mathbf{Px}}_\mathbf{j}\right)\left(\mathbf{x}_\mathbf{i}-\mathbf{x}_\mathbf{j}\right)^\mathbf{T}}\\
&=2\left(\mathbf{X}\mathbf{P}^\mathbf{T}\right)^\mathbf{T}\left[g\left(\mathbf{H}\right)+g\left(\mathbf{H}^\mathbf{T}\right)-\mathbf{H}-\mathbf{H}^\mathbf{T}\right]\mathbf{X}\\
&=2\mathbf{P}\mathbf{X}^\mathbf{T}\left(g\left(\mathbf{H}\right)-\mathbf{H}-\mathbf{H}^\mathbf{T}\right)\mathbf{X}
\end{split}
\end{equation}

\begin{figure*}[t]
	\centering
	\includegraphics[width=2.0 \columnwidth]{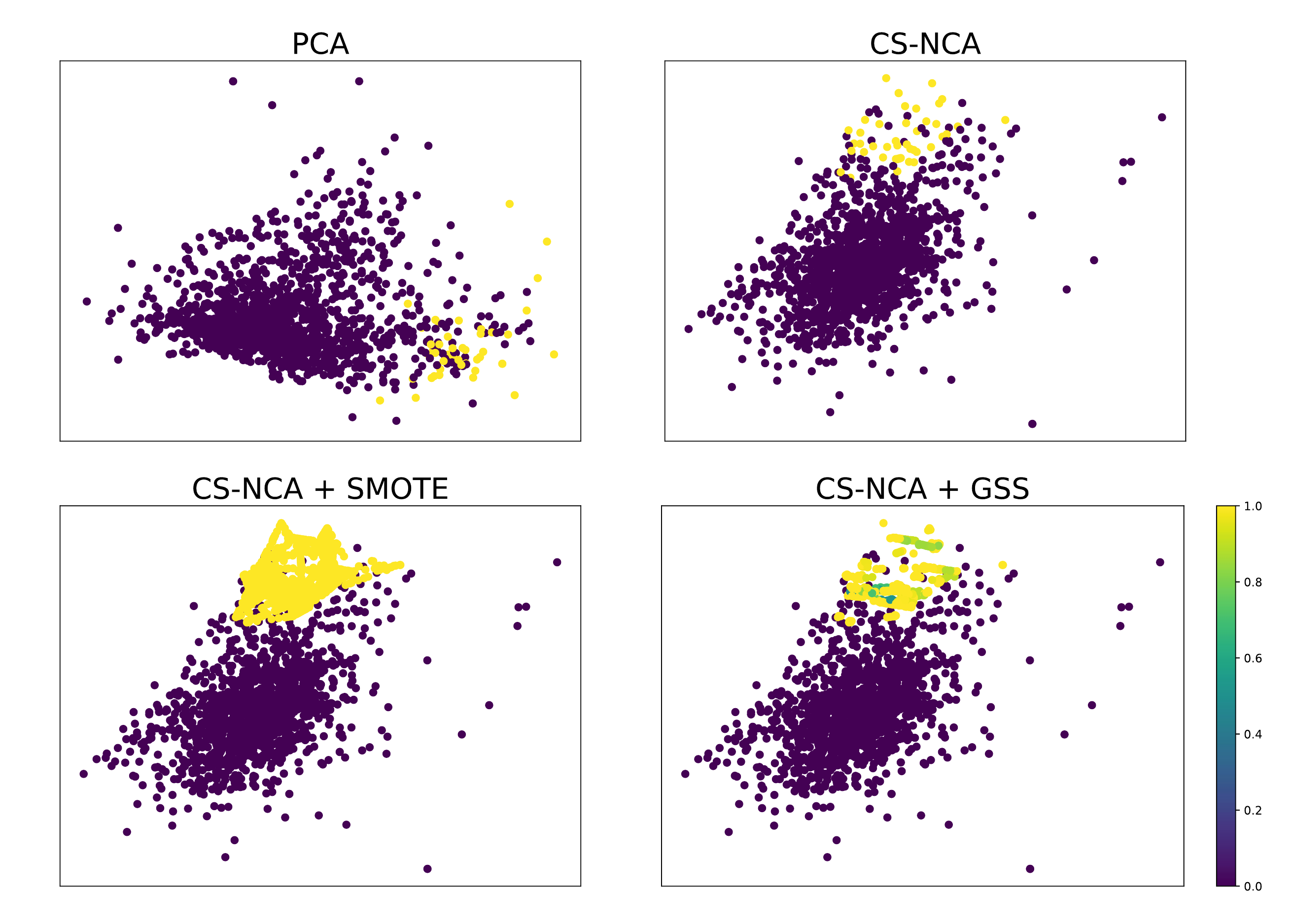}  
	\caption{Dimension reduction and oversampling in the yeast5 dataset (D=8). The above figures show the data set after PCA and CS-NCA dimensionality reduction respectively. The figures below compare the oversampling results of SMOTE and graph semi-supervised SMOTE.}
	\label{fig2}
\end{figure*}

Through the gradient descent algorithm, we can obtain the dimensionality reduction matrix $\mathbf{P}$ that maximizes the LOO accuracy of the cost sensitive nearest neighbor classifier. Fig.\ref{fig2} shows the dimensionality reduction results of principal component analysis (PCA) and CS-NCA on yeast5 dataset. The separability of CS-NCA is significantly better and the classification boundary is more distinct than that of PCA. The positive samples are more concentrated. The evidence suggests that CS-NCA can be useful for oversampling and classification.

\subsection{Graph semi-supervised SMOTE (GSS)}

\begin{figure*}[t]
	\centering
	\includegraphics[width=2.0 \columnwidth]{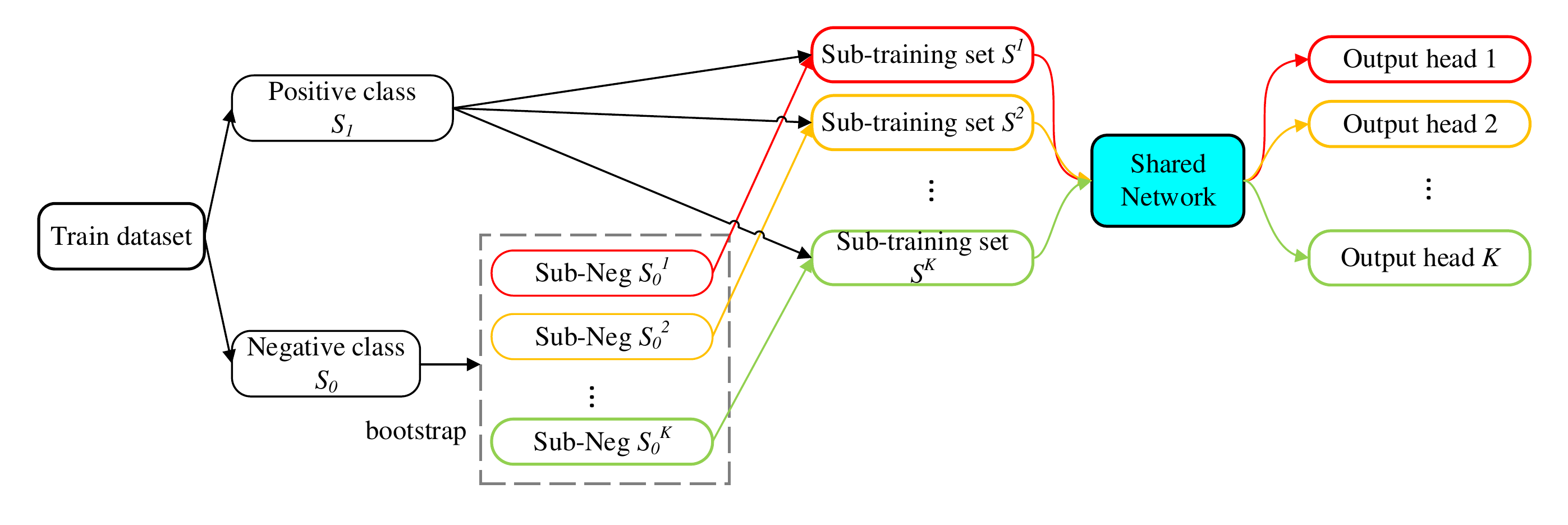}  
	\caption{Bootstrapping based ensemble framework. A single bootstrap network with $K$ output heads. Each head is trained only on its bootstrapped sub-training set.}
	\label{fig3}
\end{figure*}

For characterizing the uncertainty of the synthetic samples’ label, in this section, a graph semi-supervised algorithm \cite{RN15} is designed to relabel the synthetic samples. Given a dataset containing labeled and unlabeled samples, the idea of the graph semi-supervised algorithm is to map the dataset to a graph. Each sample corresponds to a node. If the similarity between the two samples is high, there is an edge between the corresponding nodes and the strength of the edge is proportional to the similarity between the samples. The similarity is measured by Euclidean distance, the closer the distance is, the higher the similarity is. Supposed that all labeled samples (original samples) as stained samples and all unlabeled samples (synthetic samples) as unstained samples, graph semi-supervised algorithm is similar to the process of "color" spreading on the graph. The specific process is as follows.

Assuming the training dataset after dimensionality reduction $\mathbf{S'}={\{(\mathbf{x'}_1,y_1),(\mathbf{x'}_2,y_2),...,(\mathbf{x'}_n,y_n)\}}$, the synthetic dataset generated by SMOTE is $\mathbf{S'_u}=\{\mathbf{x'}_{\mathbf{n}+\mathbf{1}},\mathbf{x'}_{\mathbf{n}+\mathbf{2}},\ldots,\mathbf{x'}_{\mathbf{n}+\mathbf{m}}\}$, where $\mathbf{x'}_\mathbf{i}\in\mathcal{X}\subseteq\mathbf{R}^d$ is the feature vector of samples, $y_i\in\mathcal{Y}=\{0,1\}$ is the label of samples, 0 is the positive (majority) class and 1 is the negative (minority) class. Construct a graph $G=(V,E)$ based on $\mathbf{S'}\cup\mathbf{S'_u}$, where the node set $V=\{\mathbf{x'}_1,\ldots,\mathbf{x'}_n,\mathbf{x'}_{n+1},\ldots,\mathbf{x'}_{n+m}\}$ and the edge set $E$ are defined as an $(n+m)\times(n+m)$ symmetric weight matrix $\mathbf{W}$. The elements in $\mathbf{W}$ are defined as:

\begin{equation}
{w_{ij}} = \left\{ \begin{array}{l}
\exp ( - \frac{\left \| {\mathbf{x}_i-\mathbf{x}_j} \right \| ^2_2}{\delta ^2} ),{\rm{\ if \ }}i \ne j\\
0,otherwise
\end{array} \right.
\end{equation}

where $\delta$ is the bandwidth parameter of Gaussian function. Thus, the nearby points in Euclidean distance are assigned large edge weight. Suppose that a real-valued function $f:V \rightarrow R$ can be learned from the graph $G=(V,E)$, $f(\mathbf{x}_\mathbf{i})$ means the probability of the sample $\mathbf{x}_\mathbf{i}$ belonging to positive class, and similar samples have similar probability. We can define the “Energy function” about $f$ as:

\begin{equation}
E\left(f\right)=\frac{1}{2}\sum_{i=1}^{n+m}\sum_{j=1}^{n+m}{w_{ij}\left(f\left(\mathbf{x'}_\mathbf{i}\right)-f(\mathbf{x'}_\mathbf{j})\right)^2}
\end{equation}

To facilitate calculation, we can rewrite the "Energy function" in matrix form. Define a diagonal matrix $\mathbf{D}=diag(d_1,d_2,\ldots,d_{n+m})$, the diagonal element is the row sum of the matrix $\mathbf{W}$, $d_i=\sum_{j=1}^{n+m}w_{ij}$. Then we can partition the matrix into blocks for labeled and unlabeled samples. 

\begin{equation}
\begin{split}
&E\left(f\right)=\mathbf{f}^\mathbf{T}\left(\mathbf{D}-\mathbf{W}\right)\mathbf{f}\\
&=\left(\mathbf{f}_\mathbf{n}^\mathbf{T}\ \mathbf{f}_\mathbf{m}^\mathbf{T}\right)^\mathbf{T}\left(\left[\begin{matrix}\mathbf{D}_{\mathbf{nn}}&\mathbf{0}_{\mathbf{nm}}\\\mathbf{0}_{\mathbf{mn}}&\mathbf{D}_{\mathbf{mm}}\\\end{matrix}\right]-\left[\begin{matrix}\mathbf{W}_{\mathbf{nn}}&\mathbf{W}_{\mathbf{nm}}\\\mathbf{W}_{\mathbf{mn}}&\mathbf{W}_{\mathbf{mm}}\\\end{matrix}\right]\right)\left[\begin{matrix}\mathbf{f}_\mathbf{n}\\\mathbf{f}_\mathbf{m}\\\end{matrix}\right]\\
&=\mathbf{f}_\mathbf{n}^\mathbf{T}\left(\mathbf{D}_{\mathbf{nn}}-\mathbf{W}_{\mathbf{nn}}\right)\mathbf{f}_\mathbf{n}-\mathbf{2}\mathbf{f}_\mathbf{m}^\mathbf{T}\mathbf{W}_{\mathbf{mn}}\mathbf{f}_\mathbf{n}\\
&\quad+\mathbf{f}_\mathbf{m}^\mathbf{T}\left(\mathbf{D}_{\mathbf{mm}}-\mathbf{W}_{\mathbf{mm}}\right)\mathbf{f}_\mathbf{m}
\end{split}
\end{equation}

where $\mathbf{f}=\left(\mathbf{f}_\mathbf{n}^\mathbf{T}\ \mathbf{f}_\mathbf{m}^\mathbf{T}\right)^\mathbf{T}$, $\mathbf{f}_\mathbf{n}^\mathbf{T}$ and $\mathbf{f}_\mathbf{m}^\mathbf{T}$ are the prediction results of function $f$ on the labeled dataset $\mathbf{S'}$ and the unlabeled dataset $\mathbf{S'_u}$, respectively. Assuming $\frac{\partial E(f)}{\partial\mathbf{f}_\mathbf{m}}=0$, we can get:

\begin{equation}
\mathbf{f}_\mathbf{m}=\left(\mathbf{D}_{\mathbf{mm}}-\mathbf{W}_{\mathbf{mm}}\right)^{-\mathbf{1}}\mathbf{W}_{\mathbf{mn}}\mathbf{f}_\mathbf{n}
\end{equation}

The function $f$ with the smallest energy satisfies $f\left(\mathbf{x'}_\mathbf{i}\right)=y_i$ on labeled samples, thus $\mathbf{f}_\mathbf{n}=\mathbf{Y}=(y_1,y_2,\ldots,y_n)$. Then we can use the obtained $\mathbf{f}_\mathbf{m}=\{f_{n+1},f_{n+2},\ldots,f_{n+m}\}$ to predict the labels of the synthetic samples. $f_i$ is the probability that the synthetic sample $\mathbf{x'}_\mathbf{i}$ belongs to positive class. The above result can also be obtained by Gaussian random fields and harmonic functions \cite{RN15}, which will not be described in detail here. 

If $f_i>p_\delta$, adding the synthetic sample $\mathbf{x}_\mathbf{i}$ into the training set, $p_\delta$ is the probability threshold to retain the synthetic sample, we set $p_\delta=0.5$. The label of new sample is two-dimensional vector $[1-f_i, f_i]$. The label of real samples are also needed to transform into one-hot vector $[0, 1]$ or $[1, 0]$. Fig.\ref{fig2} shows the oversampling results of SMOTE and GSS on yeast5 dataset. The generated samples of SMOTE fill up the space where positive samples originally exist, regardless of the distrubution of negative samples. The original classification boundary was destroyed, especially in the classes overlap area. In contrast, GSS preserves the classification boundary while oversampling, avoiding destroying the original data distribution.

\subsection{Bootstrapping based ensemble framework}
In this work, we need to solve highly imbalanced classification problem. For getting a balanced training dataset, SMOTE requires to synthesize a large number of new samples by a very limited number of positive samples. This will lead to a limited diversity of synthetic samples, which is likely to cause over-fitting problems. We propose a bootstrapping based ensemble framework (BEF), reducing the imbalanced ratio of sub-training sets while utilizing negative samples’ information  as much as possible. 

Bootstrapping is a simple technique for producing a distribution over functions with theoretical guarantees \cite{RN20}. The framework transforms the negative class set $\mathbf{S}_\mathbf{0}$ into $K$ different subsets ${\{\mathbf{S}_\mathbf{0}^\mathbf{k}\}}_{k=1}^K$ by sampling uniformly with replacement. The cardinality of subsets is equal to $IR'$ times that of the positive class set $\mathbf{S}_\mathbf{1}$, $\left|\mathbf{S}_\mathbf{0}^\mathbf{k}\right|=\ IR'\times\left|\mathbf{S}_\mathbf{1}\right|$, where $IR'$ is the imbalanced ratio of sub-training set after BEF. Combining each negative subsets with the positive set, $K$ sub-training sets ${\{\mathbf{S}^\mathbf{k}\}}_{k=1}^K$ can be obtained, $\mathbf{S}^\mathbf{k}=\mathbf{S}^\mathbf{k}_{0}\cup \mathbf{S}_{1}$. Then we train $K$ different classifiers. For each classifier $\textbf{C}_{\theta_k}$, we train the model on the dataset $\mathbf{S}^\mathbf{k}$. Each of these classifiers is trained on data from the same distribution but on a different dataset.

In cases of using neural networks as base classifier $\mathbf{C}_{\theta_k}$ , bootstrapping based ensemble framework maintains a set of $K$ neural networks ${\{\mathbf{C}_{\theta_k}\}}_{k=1}^K$ independently on $K$ different bootstrapped subsets of the data. However, it will cause the parameters of the ensemble framework to increase linearly relative to the basic classifier, and the calculation load will increase greatly \cite{RN21}. In order to remedy this issue, we adopt a single network framework which is scalable for training bootstrapped sub-training sets \cite{RN22,RN23}. The network consists of a shared network with $K$ output heads branching off independently (as shown in Fig.\ref{fig3}). Each head is trained only on its bootstrapped sub-training set. The shared network learns a joint feature representation across all the data, which can provide significant computational advantages.

Note that during training we only choose a single output head to compute the gradient and update parameters. In the test phase, we make use of all output heads to compute the weighted average result. The weight of each output head is determined by the classification result of the whole original training set $\mathbf{S}$. The weight of $i$-th output head is:

\begin{equation}
w_i=\frac{s_i}{\sum_{j=1}^{K}s_j}
\end{equation}

where $s_i$ is the classification performance of the $i$-th output head for the entire original training set $\mathbf{S}$ (in our experiment $s=F$-$measure+G$-$mean$). 
The ensemble classification result of the test sample $\mathbf{x}_i$ is:

\begin{equation}
R_i=\sum_{j=1}^{K}{{w_jo}_j(\mathbf{x}_i)}
\end{equation}

where $o_j(\mathbf{x}_i)$ is the classification result of the $j$-th output head for the test sample $\mathbf{x}_i$.

\subsection{Overall procedures of the proposed method}
Algorithm \ref{algorithmcode1} shows the overall procedures of the proposed method for imbalanced classification. Given a (two class) imbalanced dataset $\mathbf{D}$ composed of a positive (minority) class and a negative (majority) class. This imbalanced dataset is divided into training set $\mathbf{S}$ and testing set $\mathbf{T}$ based on the $n$-fold cross validation method. For training set, the first step is to perform the bootstrap-based ensemble framework to get $K$ bootstrapped sub-training sets ${\{\mathbf{S}^\mathbf{k}\}}_{k=1}^K$. Randomly select a sub-training set and use the CS-NCA algorithm to obtain the dimension reduction matrix $\mathbf{P}$. Next, for each sub-training set after dimensionality reduction, GSS is applied to obtain K balanced sub-training sets ${\{{\mathbf{S}'}^\mathbf{k}\}}_{k=1}^K$. New sub-training sets ${\mathbf{S}'}^\mathbf{i}\ (i=1,2,...K)$ is utilized to train the shared network and output head $i$. Finally, the ensemble neural network classifier is obtained. For testing set, we use all output heads of the network to obtain the weighted average result as the final prediction.

\begin{algorithm}[!t]
	\caption{Overall procedures of the proposed method}
	\label{algorithmcode1}
	\begin{algorithmic}[1]
		\Require
		training dataset, $\mathbf{S}=\{(\mathbf{x}_1,y_1),(\mathbf{x}_2,y_2),...,(\mathbf{x}_n,y_n)\}$; the number of bootstrap sets, $K$; the imbalanced ratio of sub-training set after BEF, $IR'$;\ the number of retained dimensions, $d$;
		\Ensure
		dimension reduction matrix, $\mathbf{P}$; weight vector of output head, $\mathbf{W}$;\ the ensemble neural network classifier, $\mathbf{C}_\theta$;
		\State apply the bootstrapping-based ensemble framework on $\mathbf{S}$ to get $K$ bootstrapped sub-training sets ${\{\mathbf{S}^\mathbf{k}\}}_{k=1}^K$;
		\State randomly select a sub-training set $\mathbf{S}^\mathbf{i}$;
		\State obtain the dimension reduction matrix $\mathbf{P}$ on set $\mathbf{S}^\mathbf{i}$ using CS-NCA;
		\For{$j = 1 \to K$}
		\State using the graph semi-supervised SMOTE on set $\mathbf{S}^\mathbf{j}$ after dimensionality reduction, and obtain balanced sub-training sets  ${\mathbf{S}'}^\mathbf{j}$;
		\State update the parameters of the shared network and output head $j$ by balanced sub-training sets ${\mathbf{S}'}^\mathbf{j}$;
		\State generate the classification result of the original training set $\mathbf{S}$ by the $j$-th output head, $s_j$;
		\EndFor
		\State calculate the weight vector of output heads, $\mathbf{W}$.
	\end{algorithmic}
\end{algorithm}

\section{Experiment}
In our experiments, we aim to answer three questions:

\textcircled{1} Comparing to no dimensionality reduction or other dimensionality reduction methods, can CS-NCA get a better performance?

\textcircled{2} Is the graph semi-supervised-SMOTE useful in solving imbalanced classification?

\textcircled{3} How does our approach perform on real-world datasets comparing to other benchmark methods?

To answer question \textcircled{1} and \textcircled{2}, we conduct the first experiment using 8 synthetically generated datasets. To answer question \textcircled{3}, we perform another experiment, in which we test a wide range of methods on 3 real-world datasets.

\subsection{Datasets}
\begin{figure}[t]
	\centering
	\includegraphics[width=1.0 \columnwidth]{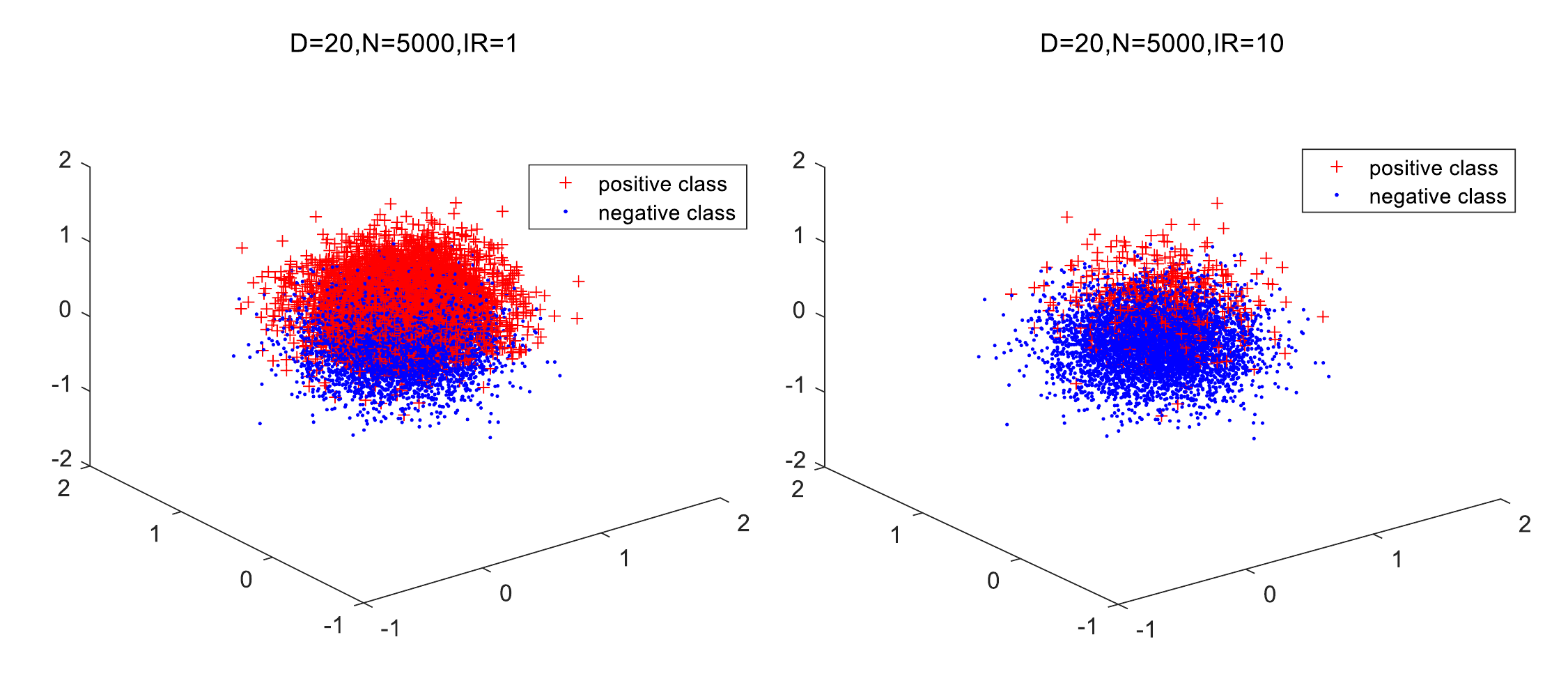}  
	\caption{Examples of synthetically generated datasets. Only the first three dimensions are shown.}
	\label{fig4}
\end{figure}

8 synthetic datasets are generated according to the method of \cite{RN24}. All of these generated datasets are binary classified and balanced. By randomly removing certain positive class samples, then datasets with different imbalanced ratios are obtained. Generating 8 datasets with different properties by setting two different levels of three parameters: (1) the dimensionality of the dataset ($D$, which is set to 20 and 100 dimensions); (2) the number of negative class samples ($n$, which is set to 1000 and 5000); (3) the imbalanced ratio of dataset ($IR$, which is set to 10:1 and 50:1). Fig.\ref{fig4} shows an example of the generated datasets.

\begin{figure*}[t]
	\centering
	\includegraphics[width=2.0 \columnwidth]{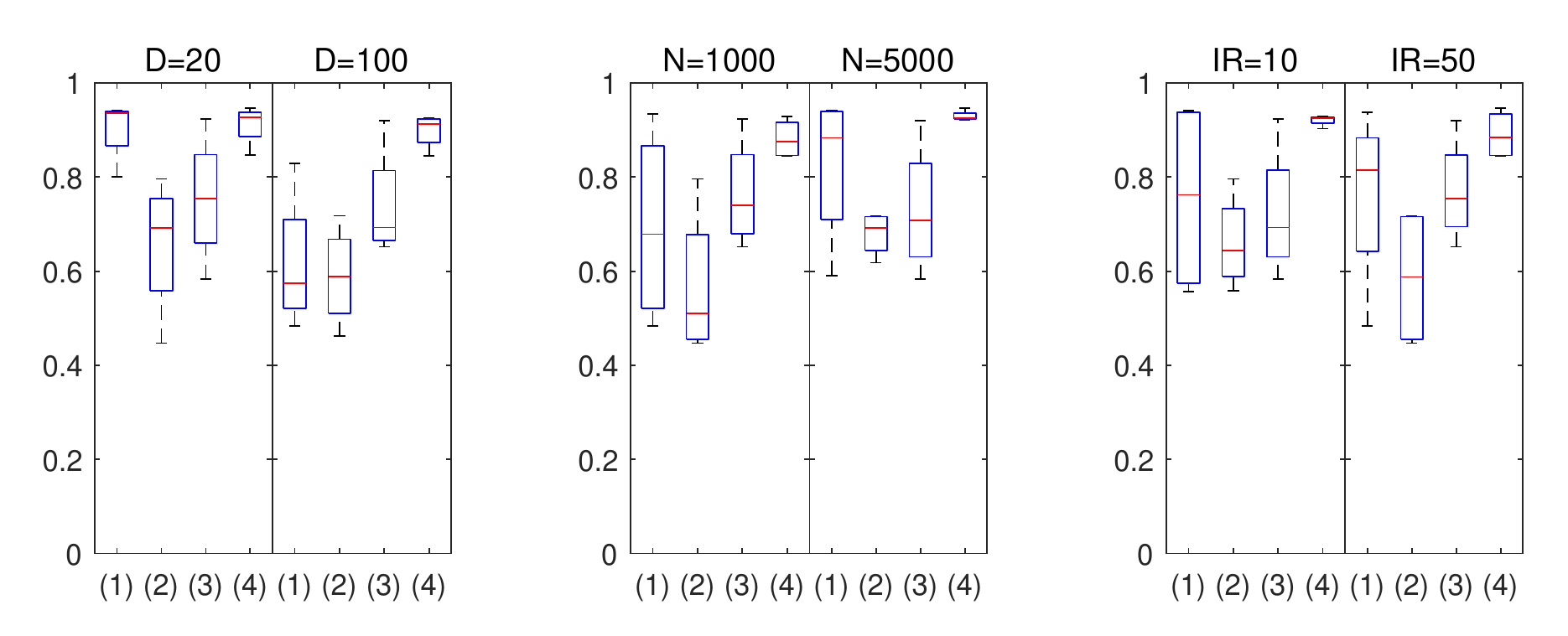}  
	\caption{Box-plots of the $G$-$mean$ for 8 synthetic datasets with different characteristics and four methods.}
	\label{fig5}
\end{figure*}

Some existing imbalanced datasets, such as the datasets in KEEL, cannot meet the experimental requirements well in terms of sample size, imbalanced ratio, and feature numbers. To better verify the effectiveness of our method, 3 representative datasets are selected from the UCI repository \cite{Dua:2019}. Datasets with different imbalanced ratios are also collected by randomly removing part of the positive samples. \textbf{Abalone} contains 4177 instances of different ages, and the number of attributes is 8. By setting abalone younger than 9 years old as negative class and others as positive class, we get the binary classification problem. Set the number of negative samples to 2000, and the number of positive samples can be obtained according to the imbalanced ratio. \textbf{Covertype} includes the forest cover type of four wilderness areas located in the Roosevelt National Forest of northern Colorado. We set Spruce-Fir as negative class and Lodgepole Pine as positive class, each sample in this dataset has 54 features and the number of negative samples is 5000. \textbf{Gisette} is one of the datasets for the NIPS 2003 variable and feature selection competition. It comes from a random subset of the “four” and “nine” patterns from the MNIST and obtained through various changes. Since the dimensions of the original dataset are too high and the information density is low, we reduce the dimension to 100 dimensions through PCA and then conduct subsequent experiments. The number of negative samples is 3500.

\subsection{Evaluation metrics}
Since the normal metric of overall accuracy in describing a imbalanced dataset classifier’s performance is no longer sufficient \cite{RN25}, another two measures $F$-$measure$ and $G$-$mean$ are used to measure the average performance of a classifier. They are given by:

\begin{eqnarray}
F\mbox{-}measure \mbox{ }&=&\mbox{ } \frac{2\cdot TP}{2\cdot TP+FN+FP} \\
G\mbox{-}mean \mbox{ }&=&\mbox{ } \sqrt{\frac{TP}{TP+FN}\cdot \frac{TN}{TN+FP}}
\end{eqnarray}

$F$-$measure$ shows the trade-off between the precision and recall regarding the positive class \cite{RN26}. $G$-$mean$ indicates the ability of a classifier to balance positive class accuracy and negative class accuracy \cite{RN27}.

To synthesize the effectiveness of $F$-$measure$ and $G$-$mean$, we take the sum of the two measures as a comprehensive evaluation metric of the classification performance, denoted as $F+G$. 

\begin{table}[t]
	\caption{Mean test ranking results for all the methods in real-world datasets with different imbalanced ratios}
	\centering
	\begin{tabular}{cccc}
		\hline
		Methods         & $F$-$measure$  & $G$-$mean$     & $F+G$          \\ \hline
		CS-SVM          & 4.455          & 2.727          & 3.182          \\
		OC-SVM          & 6.000          & 5.636          & 5.909          \\
		SMOTE-NN        & \textbf{1.182} & 3.545          & 2.545          \\
		RUS-NN          & 4.273          & 2.909          & 3.909          \\
		RUS-RF          & 2.273          & 4.909          & 3.909          \\
		Algorithm \ref{algorithmcode1} & 2.818          & \textbf{1.273} & \textbf{1.545} \\ \hline
	\end{tabular}
	\label{table1}
\end{table}

\begin{table}[t]
	\caption{Mean difference to the best result in each real-world dataset}
	\centering
	\begin{tabular}{cccc}
		\hline
		Methods         & $F$-$measure$  & $G$-$mean$     & $F+G$          \\ \hline
		CS-SVM          & 0.089          & 0.024          & 0.063          \\
		OC-SVM          & 0.173          & 0.199          & 0.323          \\
		SMOTE-NN        & \textbf{0.012} & 0.082          & 0.044          \\
		RUS-NN          & 0.099          & 0.048          & 0.096          \\
		RUS-RF          & 0.035          & 0.159          & 0.144          \\
		Algorithm \ref{algorithmcode1} & 0.057          & \textbf{0.005} & \textbf{0.012} \\ \hline
	\end{tabular}
	\label{table2}
\end{table}

\subsection{Results}
\textbf{First experiment: Synthetically generated datasets.} In this experiment, we use four approaches: (1) SMOTE-NN (SMOTE as oversampling and neural network as classifier), (2) PCA + SMOTE-NN (using PCA for dimensionality reduction), (3) CS-NCA + SMOTE-NN (using CS-NCA for dimensionality reduction) and (4) Algorithm \ref{algorithmcode1}. 

\begin{table*}[t]
	\centering
	\caption{Results of all methods in \textbf{CoverType} with different imbalanced ratios}
	\begin{tabular}{c|ccc|ccc|ccc|ccc}
		\hline
		& \multicolumn{3}{c|}{$IR=10$}                       & \multicolumn{3}{c|}{$IR=30$}                       & \multicolumn{3}{c|}{$IR=50$}                       & \multicolumn{3}{c}{$IR=100$}                      \\ \hline
		Method      & $F$\tnote{1}              & $G$\tnote{2}              & $F+G$            & $F$              & $G$              & $F+G$            & $F$              & $G$              & $F+G$            & $F$              & $G$              & $F+G$            \\ \hline
		CS-SVM      & 0.292          & 0.714          & 1.006          & 0.141          & 0.726          & 0.867          & 0.100          & 0.719          & 0.819          & 0.052          & 0.702          & 0.754          \\
		OC-SVM      & 0.174          & 0.536          & 0.710          & 0.077          & 0.551          & 0.629          & 0.051          & 0.555          & 0.606          & 0.019          & 0.481          & 0.500          \\
		SMOTE-NN    & \textbf{0.383} & 0.718          & 1.101          & \textbf{0.187} & 0.665          & 0.852          & \textbf{0.154} & 0.665          & 0.818          & \textbf{0.079} & 0.604          & 0.683          \\
		RUS-NN      & 0.326          & 0.729          & 1.055          & 0.124          & 0.686          & 0.810          & 0.103          & 0.731          & 0.834          & 0.033          & 0.601          & 0.634          \\
		RUS-RF      & 0.356          & 0.732          & 1.087          & \textbf{0.187} & 0.680          & 0.867          & 0.150          & 0.627          & 0.777          & 0.062          & 0.524          & 0.586          \\
		Algorithm \ref{algorithmcode1} & 0.352          & \textbf{0.744} & \textbf{1.096} & 0.160          & \textbf{0.741} & \textbf{0.901} & 0.119          & \textbf{0.736} & \textbf{0.855} & 0.062          & \textbf{0.719} & \textbf{0.781} \\ \hline
	\end{tabular}
	\begin{tablenotes}
		\centering
		\item[1] $F$=$F$-$measure$
		\item[2] $G$=$G$-$mean$
	\end{tablenotes}
	\label{table3}
\end{table*} 

From this experiment, we had analyzed the $G$-$mean$ for 8 generated datasets and 4 methods (a total of 32 individual results). To answer questions \textcircled{1} and \textcircled{2}, we illustrate these results under different experiment conditions in Fig.\ref{fig5}. Generally, as the dimension increases, the number of samples decreases, a degradation of classification performance can be seen. Comparing approach (1) and (2) to (3), when the dimension is high ($D$=100), the result of CS-NCA is significantly better than no dimensionality reduction or PCA. Since each dimension of the generated dataset has the same importance, when $D$=20, the performance of CS-NCA is not good compared with no dimension reduction, but it is still significantly better than PCA. All these justify the benefits of CS-NCA in processing high-dimensional imbalanced datasets. Comparing approach (3) to (4), the graph semi-supervised-SMOTE is useful in all conditions. Our proposed method obtains the best classification results on 6 out of the 8 datasets. Especially when the imbalance ratio is high, the results indicate that our proposed method makes a significant improvement compared with other methods. These results demonstrate the effectiveness of the graph semi-supervised-SMOTE to solve imbalanced classification.

\textbf{Second experiment: Real-world datasets with different imbalanced ratio.} We selected three representative real-world datasets, and randomly removed a certain number of positive samples to obtain datasets with different imbalanced ratios. We set the imbalanced ratio between 10:1 and 100:1. We choose five representative methods as benchmarks: (1) cost sensitive support vector machine (CS-SVM); (2) one class support vector machine (OC-SVM); (3) SMOTE-NN; (4) randomly undersampling and neural network as classifier (RUS-NN) and (5) randomly undersampling and random forest as classifier (RUS-RF). Because of space restrictions, we only show the test mean ranking (the lower the better) in Tab.\ref{table1}.

\begin{figure}[t]
	\centering
	\includegraphics[width=1.0 \columnwidth]{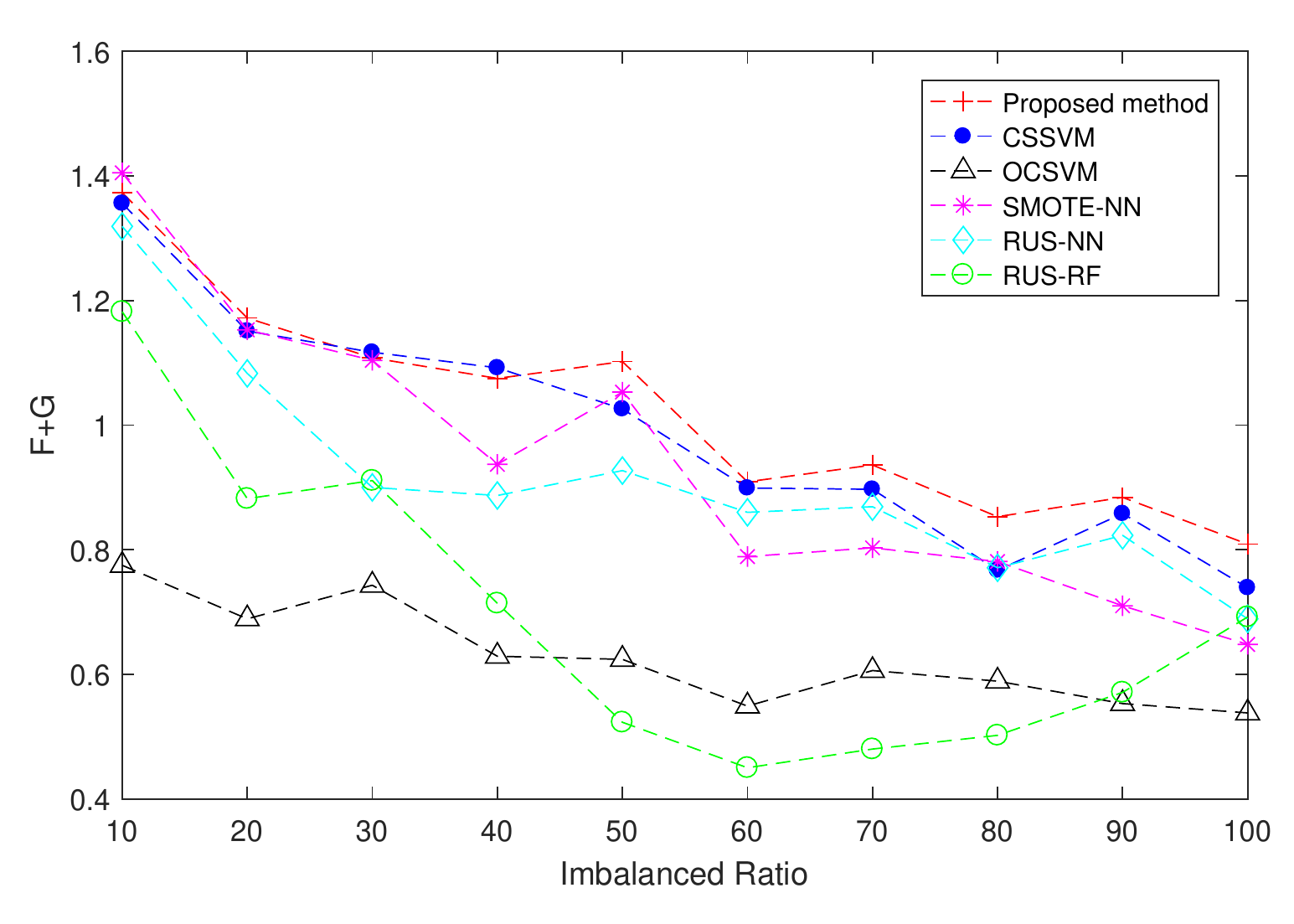}  
	\caption{Classification performance of different methods on Gisette datasets with different IR.}
	\label{fig6}
\end{figure}

Several interesting observations can be drawn. From the result of Tab.\ref{table1} and Tab.\ref{table3}, Algorithm \ref{algorithmcode1} outperforms the other five benchmark methods with $G$-$mean$ and $F+G$. In terms of $F$-$measure$, SMOTE-NN is the best among all methods. The top three methods of comprehensive performance are: Algorithm \ref{algorithmcode1}, SMOTE-NN and CS-SVM. From the details in Appendix, we can find that SMOTE-NN has good classification accuracy for negative class, but its positive sample classification accuracy is very low, often less than half. CS-SVM has the good classification accuracy for positive class, but the number of false positives (FP) is relatively high. Comparing with SMOTE-NN, the proposed method greatly improves the number of the true positives (TP). Comparing with CS-SVM, the proposed method greatly reduces the false alarm rate. The overall performance of the proposed method is the best. Although sometimes SMOTE-NN performs better than our proposed method in terms of $F$-$measure$, but its classification accuracy for positive class is less than half, which has less practical application value. Our proposed method is more suitable for high imbalance or high dimensionality datasets, and the classification performances are significantly better than others. From Tab.\ref{table2}, it is obvious that the proposed method is more robust than the benchmark methods, and the classification results are closer to the best result in each dataset regardless of the imbalanced ratio and the dimensionality.

From the results in Fig.\ref{fig6}, as the imbalanced ratio increases, the classification results also gradually deteriorate. When the imbalanced ratio is low, most methods can achieve good results. As the imbalanced ratio increases, the results of CS-SVM and the proposed method are better than other methods. When the imbalanced ratio is very high (greater than 50:1), the proposed method achieves better performance than other methods.

\section{Conclusion}
In this paper, we explore the idea of weakly supervised oversampling through graph semi-supervised learning to increase the credibility of synthetic data labels. We also propose cost-sensitive neighborhood components analysis and bootstrapping based ensemble framework to better handle the high dimensions and high imbalanced ratio problems. To verify the superiority of the proposed methods, experiments are conducted on the generated datasets and the real-world datasets. The experimental results indicate that three important components of our method are effective and practical. The proposed method outperforms other benchmarks on high imbalance and high dimensionality problems and is more robust to different imbalanced ratio. However, the classification results can still be further improved. In the future, we can design the classifier that directly deal with the label inaccurate dataset.

\bibliographystyle{IEEEtran}
\bibliography{IEEEabrv,bibfile}
\end{document}